\documentclass{ecai} 

\usepackage{latexsym}
\usepackage{amssymb}
\usepackage{amsmath}
\usepackage{amsthm}
\usepackage{booktabs}
\usepackage{enumitem}
\usepackage{graphicx}
\usepackage{color}

\usepackage{multirow}
\usepackage{soul}
\usepackage{makecell}

\newcommand{\BibTeX}{B\kern-.05em{\sc i\kern-.025em b}\kern-.08em\TeX}

\begin{document}


\begin{frontmatter}




\title{TCGPN: Temporal-Correlation Graph Pre-trained Network for Stock Forecasting}

\author[A]{\fnms{Wenbo}~\snm{Yan}}
\author[A,B,C,D]{\fnms{Ying}~\snm{Tan}}

\address[A]{School of Intelligence Science and Technology, Peking University, Beijing, China}
\address[B]{Institute for Artificial Intellignce}
\address[C]{National Key Laboratory of General Artificial Intelligence}
\address[D]{Key Laboratory of Machine Perceptron (MOE)}

\begin{abstract}
Recently, the incorporation of both temporal features and the correlation across time series has become an effective approach in time series prediction. Spatio-Temporal Graph Neural Networks (STGNNs) demonstrate good performance on many Temporal-correlation Forecasting Problem. However, when applied to tasks lacking periodicity, such as stock data prediction, the effectiveness and robustness of STGNNs are found to be unsatisfactory. And STGNNs are limited by memory savings so that cannot handle problems with a large number of nodes. In this paper, we propose a novel approach called the Temporal-Correlation Graph Pre-trained Network (TCGPN) to address these limitations. TCGPN utilize Temporal-correlation fusion encoder to get a mixed representation and pre-training method with carefully designed temporal and correlation pre-training tasks. Entire structure is independent of the number and order of nodes, so better results can be obtained through various data enhancements. And memory consumption during training can be significantly reduced through multiple sampling. Experiments are conducted on real stock market data sets CSI300 and CSI500 that exhibit minimal periodicity. We fine-tune a simple MLP in downstream tasks and achieve state-of-the-art results, validating the capability to capture more robust temporal correlation patterns.
\end{abstract}

\end{frontmatter}
\section{INTRODUCTION}
Time series data is ubiquitous in daily life, and many industries such as 
healthcare  \cite{gkotsis2017characterisation}, finance \cite{wang2021}, and transportation \cite{bui2022spatial}
healthcare, finance, and transportation generate a large amount of time series data every day. In order to discover underlying patterns and forecast future changes, utilizing these time series data for modeling has been a focal point of research across various industries. In early studies, it is commonly assumed that a time series is generated from a specific process, and the parameters of that process, such as AR \cite{yule1971method}, MA \cite{spearman1961proof}, ARMA \cite{box2015time}, ARIMA \cite{montgomery2015introduction}, ARCH \cite{engle1982autoregressive}, and GARCH \cite{engle1986modelling}, are estimated.
\begin{figure}[h]
    \centering
    \includegraphics[width=0.50\textwidth]{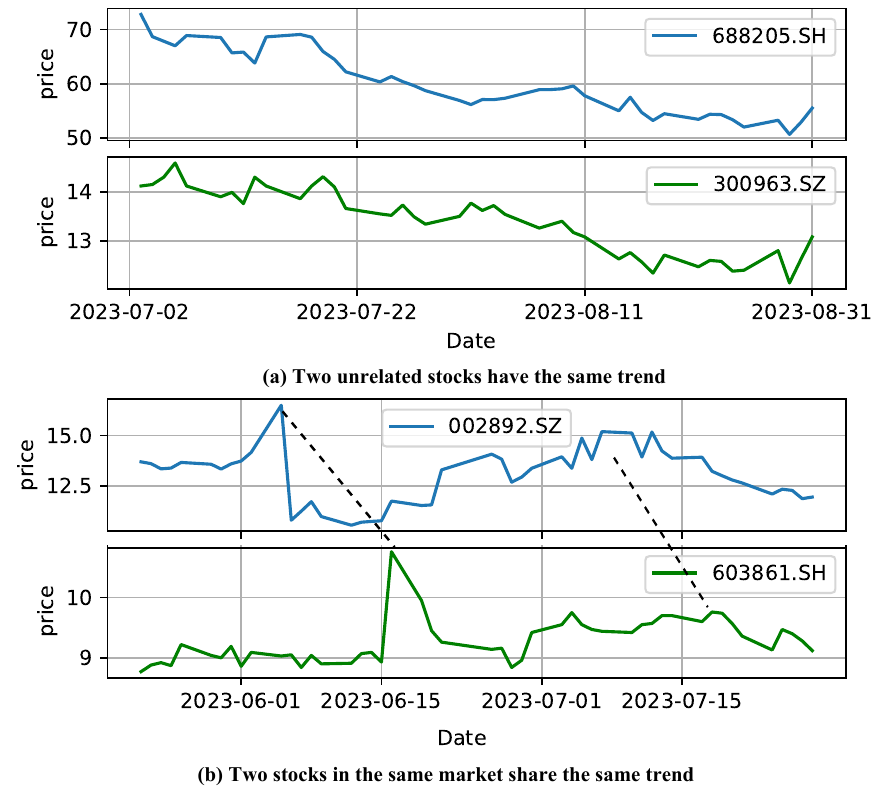}
    \caption{(a) Examples of two unrelated stocks having the same change due to similarities in time series. (b) Examples of two stock in the same industry having the same change. }
    \label{fig:fig1}
\end{figure}
In order to make more accurate predictions, many deep learning methods are applied to the field of time series forecasting in recent years. To effectively model the underlying temporal patterns,  a variety of approaches are proposed for time series modeling based on Recurrent Neural Networks (RNNs)  \cite{qin2017dual}, Time Convolutional Networks (TCNs) \cite{dai2022price}, and Transformers \cite{zhang2022crossformer}. 
In the latest research, such as  \cite{zhang2022self},  \cite{jeong2023anomalybert}, pre-training methods are introduced. At the same time, STGNNs gain increasing attention and are widely applied to time series modeling problems. STGNNs combine time series modeling with graph neural networks, effectively capturing both the internal features of time series and the temporal dependencies between them. They achieve leading results in various domains, particularly in the analysis of traffic data.

Indeed, those time series modeling methods often do not achieve satisfactory results in stock market prediction. On the one hand, stock time series often lack periodicity and exhibit non-fixed temporal patterns, requiring models to possess stronger robustness and generalization capabilities. On the other hand, there are also interdependencies among different time series, especially in stock market data. To further illustrate these characteristics, let's consider an example using real stock market data. Figure \ref{fig:fig1}(a) presents the change curves of two unrelated stocks in the same time period. It is evident that, despite belonging to different industries, based solely on the similarity of the first half of the time series, we can make similar predictions for one stock based on the changes observed in the other stock. Figure \ref{fig:fig1}(b) illustrates the change curves of two stocks from the same industry. As they belong to the same industry, the temporal patterns of the first stock appear with a delay in the changes of the second stock. This emphasizes the importance of modeling correlations in multivariate time series forecasting for stocks. At the same time, traditional STGNNs often fail to address the issue of excessive numbers of nodes. This is because the computational load for each sample typically increases quadratically with the number of nodes $N$, significantly limiting the application of STGNNs in large-scale spatio-temporal forecasting problems.

The proposed approach aims to address the aforementioned challenges by introducing a novel emporal-Correlation Graph Pre-trained
Network (TCGPN) that combines temporal features and correlation information. Pre-training allows the model to focus on the underlying patterns in the data, enhancing its robustness and generalization capabilities. The integration of time series and correlation information helps to overcome the limitations of relying solely on time series modeling, enabling a more comprehensive representation of both temporal and correlation features. The entire pre-training framework is designed to be independent of node order and quantity. It can reduce memory overhead by repeatedly sampling sub-nodes and be used on larger scale problems. 

Specifically, TCGPN first Conduct data augmentation by randomly sampling nodes, applying graph random masks, and temporal random masks, thereby exponentially generating pre-training samples. Then, it uses the Temporal-Correlation Fusion Encoder (TCFEncoder) to blend the temporal-correlation features of the data, forming an integrated encoded representation. To better explore temporal features and the correlation between sequences, TCGPN employs semi-supervised correlation tasks and self-supervised temporal tasks with designed decoder to optimize the encoded representation of the TCFEncoder. In the downstream tasks, we keep the pre-trained model frozen, and splice a simple Multi-Layer Perceptron (MLP) for specific prediction task, the parameters of the MLP can be trained. It is worth noting that the pre-trained model's output can be applied to any model or algorithm for further analysis or prediction. We compare TGCPN with STGNNs and LSTMs on two real stock datasets, demonstrating the superior performance of our model. Ablation experiments and parameter analysis are carried out to identify the optimal state of the model. In summary, the main contributions are as follows:

\begin{itemize}
    \item We propose a temporal-correlation pre-training network TCGPN which uses the Temporal-Correlation Fusion Encoder to integrate temporal and correlation features into a unified encoding, enhanced by utilizing self-supervised and semi-supervised pre-training tasks.
    \item We designed TCGPN as a structure independent of node sequence and quantity, which can exponentially increase pre-training samples through various data augmentation methods. Additionally, TCGPN can significantly reduce memory usage by repeatedly resampling when dealing with large-scale nodes.
    \item We are the first to apply TCGPN to the stock price prediction task. We conducted experiments on CSI300 and CSI500 datasets. The experimental results demonstrate that our method effectively integrates temporal and correlation patterns, achieving excellent performance in downstream prediction tasks with only simple MLP model fine-tuning.
\end{itemize}

\section{PRELIMINARY}
We first define the concepts of correlation graph and temporal-correlation forecasting problem.

Definition 2.1 \textbf{Correlation Graph} A correlation graph is used to represent the interrelationships between different time series. We use graph $\boldsymbol{G}=(\boldsymbol{V},\boldsymbol{E})$ to represent the correlation graph, where $\boldsymbol{V}$ indicates the set of $|\boldsymbol{V}|=N$ nodes and $E$ indicates the set of edges. We use adjacency matrix $\boldsymbol{A}\in{\mathbb{R}^{N \times N}}$ to represent connectivity among the graph $\boldsymbol{G}$. For each $\boldsymbol{A}_{ij}\in A$, $\boldsymbol{A}_{ij}=0$ iff $(v_i,v_j) \notin \boldsymbol{E}$ and $\boldsymbol{A}_{ij}\not=0$ iff $(v_i,v_j) \in \boldsymbol{E}$. 

Definition 2.2 \textbf{Temporal-correlation Forecasting Problem} The temporal-correlation prediction problem, also known as the spatial-temporal prediction problem, refers to the task of using historical $T$ time steps data $\boldsymbol{X}\in{\mathbb{R}^{N \times T \times F}}$ and correlation adjacency matrix $\boldsymbol{A}\in{\mathbb{R}^{N \times N}}$ to predict future values for $t$ time steps $\boldsymbol{Y}\in{\mathbb{R}^{N \times t}}$. For each sample, there are $N$ nodes, and each node has a time series $X_i$, where $X_i$ contains $T$ time steps, and each time step has $F$ features. Additionally, the correlation adjacency matrix $A$ indicates the degree of correlation between the nodes, where $a_{ij}$ represents the correlation degree between node $i$ and node $j$. The neighbors set of node $i$ is represented as $K=\{j\ |\ j \neq i \ and\  a_{ij} \neq 0\}$.

\begin{figure*}
    \centering
    \includegraphics[width=\textwidth]{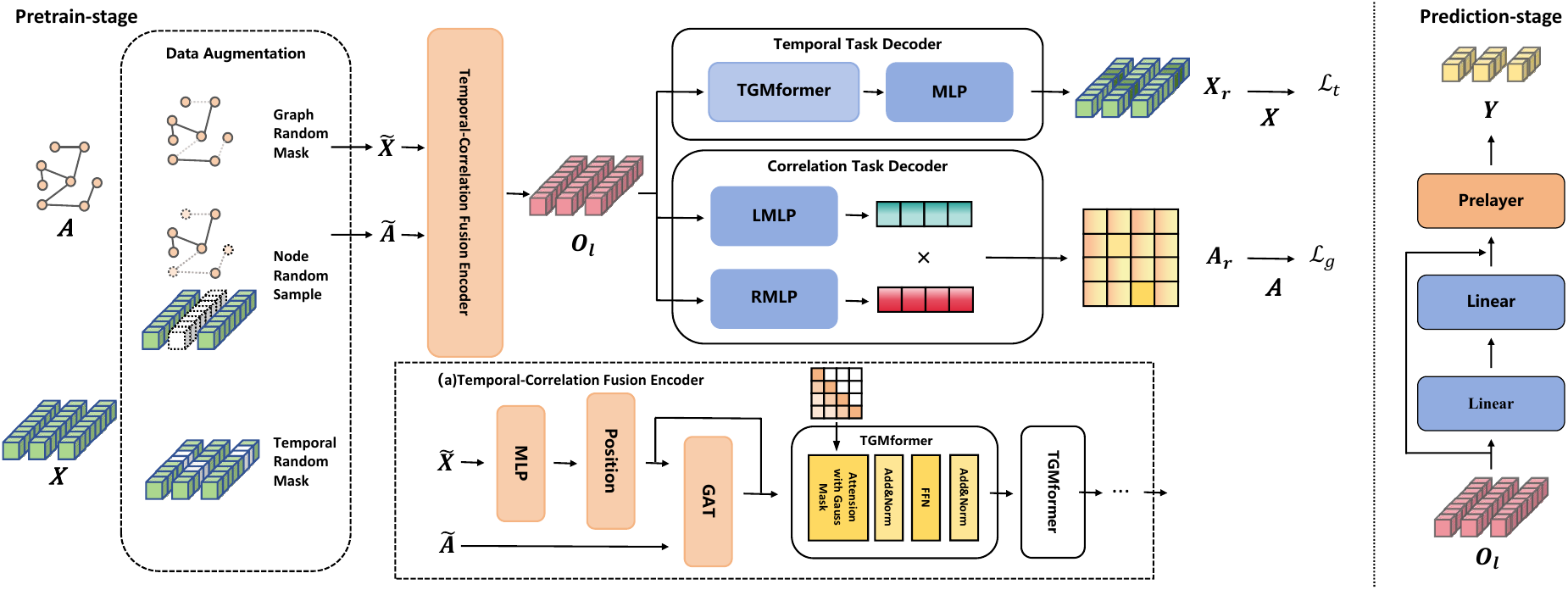}
    \caption[width=\textwidth]{The overview of the proposed temporal-correlation graph pre-trained network (TCGPN) and predict model. Left: the temporal-correlation pre-train stage. We apply multiple data augmentations to the raw data, input it into the temporal-correlation fusion encoder(TCFEncoder), and enhance model performance through self-supervised and semi-supervised tasks.
    Right: the forecasting stage. We only use the simple MLP and take the output of the TCFEncoder as the input.}
    \label{fig:enter-label}
\end{figure*}

\section{METHODOLOGY}
\subsection{Data Construction}
\noindent\textbf{Graph Definition} Graph represents the relationships between nodes, usually expressed by an adjacency matrix $\boldsymbol{A}_{N \times N}$ that indicates the strength of the correlation between any two nodes. We construct the correlation adjacency matrix from two perspectives.

(i) \textbf{Industry Graph} Taking inspiration from  \citet{HATR}, we construct the correlations based on the leading-lagging effects within the stock market industry. Industry leaders often exhibit certain changes ahead of other stocks within the same industry. Other stocks within the industry tend to follow suit with a certain time delay. We use registered capital ($R$) and turnover ($T$) as measures of industry leadership. The industry graph is a asymmetric directed graph.
\begin{eqnarray}
   \begin{cases}
        a_{ij}=\frac{R_j}{R_i}+\frac{T_j}{T_i},\ i,j\ in\ the\ same\ industry\\
        a_{ij}=0,\ else
    \end{cases} 
\end{eqnarray}

(ii) \textbf{Distance Graph} The Industry graph constructs correlations suitable for stock data based on prior knowledge, but this undoubtedly introduces additional information, which is unfair for some STGNNs. Therefore, we define another correlation between two time series by utilizing the Euclidean distance between time series. The distance graph is a symmetric correlation graph.
\begin{eqnarray}
   a_{ij}=||\boldsymbol{X_i}-\boldsymbol{X_j}||_2 
\end{eqnarray}
\noindent\textbf{Data Augmentation}
As mentioned above, our model can input an arbitrary number of nodes in any order, which will be described in detail later. Therefore, we can enhance the data in various ways to achieve better pre-training effects, and to have stronger modeling capabilities and robustness for different data.

(i) \textbf{Node Random Sampling} Since TCGPN is node-free, we can randomly sample a variety of training samples within a certain range of node counts. On one hand, this approach can generate a large number of training samples on the original dataset, enhancing the model's pre-training effect. On the other hand, for spatio-temporal prediction problems with a particularly large number of nodes, the training memory consumption can be greatly reduced through multiple sampling, achieving an effect similar to training with all nodes together. For each training sample containing $N$ nodes, there is a time series $\boldsymbol{X} \in \mathbb{R}^{N \times T \times F}$ and an adjacency matrix $A \in \mathbb{R}^{N \times N}$, which is calculated from an industry graph or a distance graph. 

(ii) \textbf{Graph Random Mask} According to the needs of the semi-supervised pre-training task, we randomly mask nodes in the adjacency matrix at a certain graph masking rate $r_g$. By masking multiple times, we can exponentially increase the pre-training samples, better enhancing the model's performance. For the masked adjacency matrix $\boldsymbol{\widetilde{A}}$, we normalize the non-zero nodes of the adjacency matrix as one of the inputs for the subsequent model.
\begin{eqnarray}
    \boldsymbol{\widetilde{A}}=Norm(Mask(\boldsymbol{A}))
\end{eqnarray}

(iii) \textbf{Temporal Random Mask} Similar to graph random mask, for the needs of self-supervised pre-training, we mask the time series at a temporal masking rate of $r_t$ and obtain the mask time series $\boldsymbol{\widetilde{X}}$. It's worth noting that, different from randomly selecting time steps, inspired by \citet{STEP},we randomly select a starting time step  and mask a continuous period of time after it. This approach can enhance the challenge of the task and exponentially increase the samples to achieve better results.

\subsection{Temporal-Correlation Fusion Encoder}
To obtain the spatio-temporal integrated encoding, we designed Temporal-Correlation Fusion Encoder (TCFEncoder), which includes three parts: position encoding, correlation fusion and temporal encoding.

For the initial fusion of features and to ensure not lost the temporal sequence order in subsequent encoding processes, we add position encoding \citep{vaswani2017attention} to the input time series $\boldsymbol{\widetilde{X}}$
\begin{eqnarray}
    \boldsymbol{\hat{X}}=Position(W^T_{f}\boldsymbol{\widetilde{X}}+b)
\end{eqnarray}
where $Position$ is sin/cos positional encoding influenced by the time step $t$ and feature position $f$ \citep{vaswani2017attention}, $W$ is the feature fusion weights and $b$ is the bias. After this step, the features of the nodes are preliminarily integrated and mapped to a higher dimensional space, and the corresponding positional information is introduced for each variable at every time step.

For graphs constructed with some prior knowledge, they are undoubtedly incomplete and biased. To effectively fuse correlation information, we apply Graph Attention Network (GAT)  \citep{velivckovic2017graph} into the structure. GAT has a learnable attention mechanism, where the relevance between nodes is adaptively calculated by the attention mechanism:
\begin{eqnarray}
    \alpha_{ij}=\frac{exp(\sigma(\boldsymbol{\vec{a}^T}[\boldsymbol{W}\widetilde{x}_i||\boldsymbol{W}\widetilde{x}_j]))}{\sum_{k\in N_i}{ exp(\sigma(\boldsymbol{\vec{a}^T}
[\boldsymbol{W}\widetilde{x}_i||\widetilde{W}\widetilde{x}_j]))}}
\end{eqnarray}
where $\boldsymbol{W} \in \mathbb{R}^{F \times F^`}$ is the weight matrix, $\boldsymbol{\vec{a}^T}$ is a single-layer feed-forward neural network, and the $\sigma$ is LeakyReLU activate function. Based on the relevance $\alpha_{ij}$ between nodes and the neighbor set $N$ obtained from the adjacency matrix $\boldsymbol{A}$, we can encode the correlation features into the time series of each node:
\begin{eqnarray}
    z_i=\sigma (\frac{1}{K} \sum_{K=1}^{K}\sum_{j \in N_i}\alpha_{ij}\boldsymbol{W}\vec{x}_j)
\end{eqnarray}
where $N_i$ is the set of neighbor node of node $i$, and $K$ is the number of head of attension. The collection of all time series can be represented as $\boldsymbol{Z}$.

To explore the temporal series features on the basis of mixing correlation information, a transformer structure with a temporal gaussian mask, called TGMformer, is utilized
\begin{eqnarray}
    \begin{split}
        \boldsymbol{Q}=\boldsymbol{W_Q^T}\boldsymbol{Z},\boldsymbol{K}=\boldsymbol{W_K^T}\boldsymbol{Z},\boldsymbol{V}=\boldsymbol{W_V^T}\boldsymbol{Z}\\
        \boldsymbol{O_l}=softmax(\frac{\boldsymbol{Q}\boldsymbol{K^T}}{\sqrt{d_k}}\cdot \boldsymbol{M})\boldsymbol{V}
    \end{split}
\end{eqnarray}
where  $\boldsymbol{W_Q}$, $\boldsymbol{W_K}$, $\boldsymbol{W_V}$ are independent weights matrix that map $\boldsymbol{Z}$ to three different spaces, $O$ is the encoding with temporal-correlation feature fused. and $\boldsymbol{M}$ is temporal gaussian mask calculated by
\begin{eqnarray}
    m_{ij}=
    \begin{cases}
    0, \ j>i\\
    exp(-\frac{(j-i)^2}{2[\sigma_h]^2}),\ else 
    \end{cases}
\end{eqnarray}
where $\sigma_h$ is the prior controlling the concentration of Gaussian distribution.Introducing the temporal Gaussian mask can, on one hand, prevent historical time steps from observing future information, emphasizing the impact of historical features at different time scales. On the other hand, it can effectively model the time decay effect of historical influences, as mentioned by  \citet{ding2020hierarchical}.

It is worth emphasizing that the correlation coefficients during the encoder process are calculated by GAT based on time series, and TGMformer encodes each time series independently and irrespective of node order. Therefore, our model is invariant to node order, and can enhance pre-training performance through various data augmentation methods.

\subsection{Temporal Self-Supervised Task}
We design temporal self-supervised tasks to enhance the temporal encoding ability and temporal-correlation fution ability. The basic idea is that the missing time steps can be inferred from the context of the time series on one hand, and on the other hand, can be deduced from similar series based on the correlation between time series. Therefore, an effective temporal-correlation fusion encoding should be able to recover the missing time steps to a certain extent.

Specifically, we decode the output $\boldsymbol{O_l}$ of the Temporal-Correlation Fusion Encoder, using a TGMformer with future information masked, as the decoder. The output of the decoder is then mapped through a fully connection layer (FC) to the same shape as the unmasked time series $\boldsymbol{X}$. 

\begin{eqnarray}
    \boldsymbol{X_r}=FC( TGMformer(\boldsymbol{O_l}) )
\end{eqnarray}

We optimize the mean squared error loss of the masked parts to make the recovered data closer to the real data.
\begin{eqnarray}
    \mathcal{L}_{t} =\frac{1}{N}\sum_{N}||(\boldsymbol{X}-\boldsymbol{X_r})\cdot \widetilde M_{t}||
\end{eqnarray}
where $\widetilde M_{t}$ means calculate loss only on the masked position. 

Under the influence of this self-supervised task, the more accurate the recovery of temporal data is, it indicates that the temporal-correlation fusion encoder can more effectively explore the latent temporal patterns in the variations of time series and make effective inferences about the missing time steps. At the same time, it can learn better ways of integrating information owned by the neighbor nodes, more accurately assess the correlation with neighbor nodes, and integrate similar time series features into its own encoding.

\subsection{Correlation Semi-Supervised Task}
For spatio-temporal forecasting issues, the correlation between nodes is often not directly assessable. To ensure the effectiveness of the correlation while breaking through the limitations of prior knowledge, we have meticulously designed a semi-supervised task for correlation. We aim to adaptively learn the degree of correlation between nodes under the guidance and constraints of a predefined graph, uncovering more effective correlations while incorporating prior knowledge, and achieving stronger generalizability and robustness across all nodes.

Specifically, our predefined graph calculates the degree of correlation between nodes based on prior knowledge, but in the temporal-correlation fusion encoder, GAT only uses the connectivity between nodes. The degree of correlation between nodes is adaptively learned and ultimately integrated into the fusion encoding $O_l$. We designed a key-value structure adjacency matrix decoder to recover an adjacency matrix $\boldsymbol{A_r}$from node encodings. 
\begin{eqnarray}
    \begin{aligned}
        \boldsymbol{L}=\boldsymbol{W_L^TT}+b_L\\
        \boldsymbol{R}=\boldsymbol{W_R^TT}+b_R\\
        \boldsymbol{\widehat{A}}=\boldsymbol{L}\boldsymbol{R^T}
    \end{aligned}
\end{eqnarray}
where $\boldsymbol{W_L}$, $\boldsymbol{W_R}$ are independent weights matrix, $b_L$,$b_R$ are independent bias matrix. In data augmentation, we mentioned that the adjacency matrix of the sample is randomly masked. We use the unmasked parts as supervision for the learned degree of correlation, ensuring that the adaptively computed results do not deviate excessively from prior knowledge, while maintaining the nodes' generalizability in other positions.
\begin{eqnarray}
    \mathcal{L}_{g} =\frac{1}{N}\sum_{N}||(\boldsymbol{A}-\boldsymbol{A_r})\cdot M_{g}||
\end{eqnarray}
where $M_{g}$ means calculate loss only on the unmasked position. What`s more, by controlling the masking rate $r_{g}$, we can effectively balance the prior knowledge and generalization, achieving a trade-off in correlation relationships.
\subsection{Fine-tune for Prediction}
Due to our belief that the temporal features and correlation of time series have been effectively fused together in the pre-training process, forming a comprehensive latent feature representation $\boldsymbol{O}$, we only utilize a simple MLP for fine-tuning on downstream tasks.

During the prediction phase, we no longer apply masking to the inputs of the temporal and correlation graphs. Similarly, we no longer use the temporal decoder and correlation decoder from the pre-trained model. We directly input the complete time series and adjacency matrix into the pre-training model, and use the output $\boldsymbol{O_l}$ from the encoder as the input for the downstream model. The model structure can be represented as follows.
\begin{eqnarray}\boldsymbol{O_1}=fc(relu(fc(\boldsymbol{O_l}))+\boldsymbol{O_l}\end{eqnarray}
\begin{eqnarray}\boldsymbol{O_2}=predict\ layer(\boldsymbol{O_1})\end{eqnarray}

The comprehensive latent feature representation $\boldsymbol{O_l}$ is first passed through two linear layers to obtain a high-dimensional feature representation, followed by a ReLU non-linear activation function. The output of the linear layers is then combined with the initial comprehensive features through a residual connection, preserving the initial aggregated features. Subsequently, the predict layer aggregates information from multiple time steps and maps the features to the desired output dimension. With simple modifications, we obtain the output $\boldsymbol{\widehat{Y}}$ for a specific prediction task.

To achieve good performance on time series prediction tasks, we consider both the accuracy of the data itself and the correlation of predictions across time steps in the loss function of the downstream task. First, we use MSE loss as the loss function to measure the accuracy of the predictions.

\begin{equation} \mathcal{L}_{mse} =\frac{1}{N}\sum_{N}{(\widehat{y_i},y_i)}\end{equation}

Additionally, we use the Pearson correlation coefficient to measure the correlation of predictions within a time step 
\begin{equation} \mathcal{L}_{pearson} =-\frac{\sum_{i=1}^{N}(y_i-\overline{y})(\widehat{y_i}-\overline{\widehat{y}})}{\sqrt{\sum_{i=1}^{N}(y_i-\overline{y})}\sqrt{\sum_{i=1}^{N}(\widehat{y_i}-\overline{\widehat{y}})}}\end{equation}
The ultimate form of loss function is as follows
\begin{equation}\mathcal{L}=\lambda_{m}\mathcal{L}_{mse}+\mathcal{L}_{person}\end{equation}
where $\lambda_m$ is the weight of $\mathcal{L}_{mse}$.

\section{EXPERIMENTS}
In this section, we conduct experiments on two real stock market datasets to demonstrate the effectiveness of our time-correlation pre-training method. We aim to show that it can replace existing STGNNs on non-periodic time series data. Additionally, we perform a comprehensive evaluation of the implementation, assessing the impact of various components and parameters on the experimental results.
\begin{table}[hb]   
    \centering
    \resizebox{\linewidth}{!}{

    \begin{tabular}{c|c|c|c|c|c}
        \hline
        $\textbf{Dataset}$ & $\textbf{Samples}$ & $\textbf{Node}$ & $\textbf{Sample Rate}$ & $\textbf{Time Span}$ &\textbf{ Partition}\\
        \hline
        CSI300& 3159 & 300 & daily & 12years& 10/1/1\\ 
        CSI500& 3159 & 500 & daily & 12years& 10/1/1\\ 
        \hline
    \end{tabular}
    }
    \caption{The overall information for datasets}
    \label{tab:dataset}
\end{table}
\subsection{Experimental Setup}
\noindent\textbf{Datasets}
We conduct detailed experiments on two real-world stock datasets:
\begin{itemize}
    \item CSI300: CSI300 is a stock dataset that contains the performance of the top 300 companies listed on the Shanghai and Shenzhen stock exchanges in China. It contains daily frequency data for 300 stocks from 2010 to 2022, with a total time step of 3159 and a feature number of 45.
    \item CSI500: CSI500, is a stock dataset that contains the performance of 500 small and medium-sized companies listed on the Shanghai and Shenzhen stock exchanges. It contains daily frequency data for 500 stocks from 2010 to 2022, with a total time step of 3159 and a feature number of 45.
\end{itemize}

Brief statistical information is listed in Table \ref{tab:dataset}. Considering the requirements for pre-training data volume, we divided the data by year. We used ten years of data as the training set, one year of data as the validation set, and one year of data as the test set.

\noindent\textbf{Baseline}
For the sake of fair comparison, we divided the experiment into two baseline groups based on whether additional data was used.

Temporal only group: we used time series and distance-based graphs as inputs without introducing any additional information. We employed a temporal neural network TPA-LSTM  \cite{TPALSTM}, and six spatial-temporal graph neural networks ASTGCN \cite{ASTGCN}, MSTGCN \cite{MSTGCN}, MTGNN \cite{MTGNN}, STEMGNN \cite{stemGNN}, STSGCN \cite{STSGCN}, STGCN \cite{STGCN}.

Industry graph group: we utilized time series and pre-defined industry graphs as inputs, which introduced additional information compared to distance graph. We compared four spatial-temporal graph neural networks ASTGCN \cite{ASTGCN}, MSTGCN \cite{MSTGCN}, STEMGNN \cite{stemGNN}, STSGCN \cite{STSGCN}, STGCN \cite{STGCN} which allowed pre-defined graph structures.

\noindent\textbf{Metrics}
We evaluate the performances of all baseline by nine metrics which are commonly used in stock prediction task including Information Coefficient (IC), Profits and Losses (PNL), Annual Return (AR), Volatility (VOL), Sharpe Ratio (Sharpe), Max Drawdown (MDD), Calmar Ratio (Calmar), Win Rate (WinR), Profit/Loss Ratio (PL-ratio).

\noindent\textbf{Implementation} We set the time step to 30 and form cross-sections of all time series within the same time interval as inputs to the model. We then calculate the correlation graph $G$ accordingly. The masking rate for the time series $r_t$ is 0.3, and the masking rate for the adjacency matrix $r_g$ is 0.3 also. For the Graph Attention Network, we set the number of heads to 4 and the output dimension to 32. For TGMformer, we configure the Encoder with 2 TMBlocks for the CSI300 task and 3 TMBlocks for the CSI500 task. The number of temporal heads for both tasks is 8, and the output dimension is 128. The Decoder consists of only one TGMformer layer. Considering the range of values for the two losses, $L_{pearson}$ is in [-0.2,0], while that of $L_{MSE}$ is in [0, 1]. $\lambda_m$ is set to 0.3 to bring the values of the two loss functions into the same range.

\begin{table*}[htbp]
  \centering
  \resizebox{\linewidth}{!}{
    \normalsize
    \renewcommand{\arraystretch}{1.1}
    \begin{tabular}{cc|ccccccccc}
    \hline
    \hline
    \textbf{Datasets} & \textbf{Methods} & \multicolumn{1}{c}{IC} & \multicolumn{1}{c}{PNL} & \multicolumn{1}{c}{AR} & \multicolumn{1}{c}{VOL} & \multicolumn{1}{c}{MDD$\downarrow$} & \multicolumn{1}{c}{Sharpe} & \multicolumn{1}{c}{Calmar} & \multicolumn{1}{c}{WinR} & \multicolumn{1}{c}{PL-ratio} \\
    \hline
    \hline
    \multicolumn{1}{c}{\multirow{8}[3]{*}{\textbf{\makecell[c]{CSI300 \& \\Distance Graph}}}} & \textbf{ASTGCN} & \ul{0.07094}  & 0.21611  & 0.21433  & 0.10050  & 0.22884  & \ul{2.13256}  & \ul{0.93659}  & 0.52479  & 1.43542  \\
          & \textbf{MSTGCN} & 0.04937  & 0.05839  & 0.05791  & 0.09311  & \ul{0.22044}  & 0.62195  & 0.26270  & 0.43388  & 1.11069  \\
          & \textbf{MTGNN} & 0.06379  & \ul{0.24032}  & \ul{0.24808}  & 0.09684  & 0.23243  & 1.99332  & 0.68244  & 0.54339  & \textbf{1.59912 } \\
          & \textbf{STEMGNN} & 0.04813  & 0.13201  & 0.13092  & 0.08169  & \textbf{0.18127 } & 1.60262  & 0.72223  & 0.51653  & 1.29904  \\
          & \textbf{TPA-LSTM} & 0.04391  & 0.20293  & 0.20125  & \ul{0.10110}  & 0.29634  & 1.99062  & 0.67913  & \textbf{0.57025 } & 1.37415  \\
          & \textbf{STGCN} & 0.05891  & 0.15418  & 0.15092  & 0.08467  & 0.32435  & 1.41439  & 0.40459  & 0.53512  & 1.49597  \\
          & \textbf{STSGCN} & 0.03002  & 0.05020  & 0.04978  & 0.08395  & 0.54290  & 0.82836  & 0.71194  & 0.46256  & 1.08462  \\
\cmidrule{2-11}          & \textbf{TCGPN} & \textbf{0.09051 } & \textbf{0.26305 } & \textbf{0.26088 } & \textbf{0.10131 } & 0.22467  & \textbf{2.57502 } & \textbf{1.16115 } & \ul{0.54959}  & \ul{1.50315}  \\
    \hline
    
    \multicolumn{1}{c}{\multirow{8}[3]{*}{\textbf{\makecell[c]{CSI500\&\\DistanceGraph}}}} & \textbf{ASTGCN} & \ul{0.19006}  & \ul{1.21366}  & \ul{1.20363}  & 0.13580  & 0.14789  & 8.79827  & 8.13862  & \ul{0.71488}  & 4.39228  \\
          & \textbf{MSTGCN} & 0.14615  & 1.19932  & 1.18941  & 0.11405  & \textbf{0.11114 } & \textbf{9.90762 } & \textbf{10.16688 } & 0.71446  & \textbf{4.61038}  \\
          & \textbf{MTGNN} & 0.16523  & 1.02346  & 1.01500  & 0.13533  & 0.18518  & 7.28482  & 5.48108  & 0.68182  & 3.65099  \\
          & \textbf{STEMGNN} & 0.17148  & 1.10276  & 1.09365  & 0.12169  & \ul{0.11500}  & 8.88748  & 8.51028  & 0.71074  & 4.29236  \\
          & \textbf{TPA-LSTM} & 0.12073  & 1.21024  & 1.20024  & \textbf{0.14002 } & 0.12534  & 8.22924  & \ul{9.19302}  & 0.66645  & 4.45730  \\
          & \textbf{STGCN} & 0.07288  & 0.62057  & 0.61544  & 0.10853  & 0.14981  & 4.74364  & 4.93818  & 0.61364  & 2.89669  \\
          & \textbf{STSGCN} & 0.06873  & 0.52623  & 0.52188  & 0.09195  & 0.36341  & 4.67015  & 4.32179  & 0.57645  & 3.60648  \\
\cmidrule{2-11}          & \textbf{TCGPN} & \textbf{0.19619 } & \textbf{1.22877 } & \textbf{1.21862 } &\ul{0.13655}  & 0.16644  & \ul{8.92447}  & 8.32171  & \textbf{0.73967 } & \ul{4.49383 } \\
    \hline
    \hline
    \end{tabular}%
    
    }
      \caption{Results of temporal only group. Compared to STGNNs and LSTMs on stock time series and a distance graph built from temporal data. $\downarrow$ indicates that the smaller the metric is better. The best result is in bold, and the second result is underlined.}
  \label{tab:main result}%
\end{table*}%

\begin{table*}[h]
  \centering
    \resizebox{\linewidth}{!}{
        \normalsize
    \renewcommand{\arraystretch}{1.1}
    \begin{tabular}{c||cccccc|cccccc}
    \multicolumn{1}{r}{} &       &       &       &       &       & \multicolumn{1}{r}{} &       &       &       &       &       &  \\
\hline
\hline
    \multirow{2}[4]{*}{\textbf{Metrix}} & \multicolumn{6}{c|}{\textbf{CSI300 \& Industry Graph}} & \multicolumn{6}{c}{\textbf{CSI500 \& Industry Graph}} \\

 & \multicolumn{1}{c}{\textbf{TCGPN}} & \multicolumn{1}{c}{\textbf{ASTGCN}} & \multicolumn{1}{c}{\textbf{MSTGCN}} & \multicolumn{1}{c}{\textbf{STEMGNN}} & \multicolumn{1}{c}{\textbf{STGCN}} & \multicolumn{1}{c|}{\textbf{STSGCN}} & \multicolumn{1}{c}{\textbf{TCGPN}} & \multicolumn{1}{c}{\textbf{ASTGCN}} & \multicolumn{1}{c}{\textbf{MSTGCN}} & \multicolumn{1}{c}{\textbf{STEMGNN}} & \multicolumn{1}{c}{\textbf{STGCN}} & \multicolumn{1}{c}{\textbf{STSGCN}} \\
\hline
\hline
    IC    & \textbf{0.10548 } & \ul{0.08142}  & 0.04483  & 0.04133  & 0.06313  & 0.04308  & \textbf{0.19352 } & \ul{0.18838}  & 0.15670  & 0.18422  & 0.15893  & 0.17029  \\
    PNL   & \textbf{0.49322 } & \ul{0.29625}  & 0.19968  & 0.22040  & 0.24797  & 0.21004  & \textbf{1.22555 } & 0.46641  & \ul{1.16562}  & 1.04978  & 0.81531  & 0.87598  \\
    AR & \textbf{0.48914 } & \ul{0.29381}  & 0.19803  & 0.21858  & 0.24592  & 0.20830  & \textbf{1.21542 } & 0.46255  & \ul{1.15598}  & 1.04111  & 0.81353  & 0.86874  \\
    VOL & \ul{0.09353}  & \textbf{0.09949 } & 0.08860  & 0.08649  & 0.08404  & 0.08754  & \ul{0.12754}  & 0.12657  & \textbf{0.13311 } & 0.11832  & 0.10547  & 0.10186  \\
    MDD$\downarrow$ & \textbf{0.12525 } & 0.25073  & \ul{0.17240} & 0.17969  & 0.21157  & 0.17605  & \ul{0.08159}  & 0.17642  & 0.13470  & \textbf{0.08015 } & 0.33949  & 0.36223  \\
    Sharpe & \textbf{5.23004 } &\ul{2.95314}  & 2.23518  & 2.52716  & 2.59416  & 2.38117  & \textbf{9.52982 } & 3.65458  & 8.68429  & \ul{8.79931}  & 7.02448  & 5.72384  \\
    Calmar & \textbf{3.90532 } & 1.17179  & 1.14863  & \ul{1.21644}  & 1.16021  & 1.18254  & \textbf{14.89728 } & 2.62191  & 8.58164  & \ul{12.98917}  & 10.62898  & 9.88350  \\
    WinR & \textbf{0.61570 } & 0.53719  & 0.52479  & \ul{0.54232}  & 0.53099  & 0.53356  & \textbf{0.69008 } & 0.55785  & 0.66529  & \ul{0.68595}  & 0.54215  & 0.56322  \\
    PL-ratio & \textbf{2.37183 } & \ul{1.63736}  & 1.44192  & 1.49903  & 1.53964  & 1.47048  & \textbf{5.39801 } & 1.82313  & \ul{4.77778}  & 4.69771  & 3.71113  & 3.34450  \\
\hline
\hline
    \end{tabular}%
    }    
    \caption{Results of industry graph group. Compared to STGNNs on stock time series and a industry graph built from market data. $\downarrow$ indicates that the smaller the metric is better. The best result is in bold, and the second result is underlined.}
  \label{tab:result indus group}%

\end{table*}%

\subsection{Main Results}
Table \ref{tab:main result} and Table \ref{tab:result indus group} summarize the evaluation results of temporal only group and industry graph group. The best results are highlighted in bold font and the second results are highlighted by underline. In summary, TCGPN achieves state-of-the-art preformance in both two tasks.

\noindent\textbf{Temporal only group} In this task, we compare our TCGPN with other Spatial-Temporal Graph Neural Networks and Long Short Term Memory Models (LSTMs). Table \ref{tab:main result} shows the detailed experimental results on dataset with Distance Graph. Our model achieves the best results on multiple metrics across two datasets. Especially for IC, PNL and AR, our model achieves the best effect on both datasets and surpass the second place by a large margin. For VOL, MDD, Sharpe, Calmar, WinR and PL-ratio, we can steadily achieve the first and second results. In general, our model can achieve the best or second best results on the vast majority of metrics, while other baseline models, whether STGNNs or LSTMs, even achieve the best results on individual metric, but at the same time other metrics cannot compete with our TCGPN. This is because the setting of our model structure and pre-training tasks pay more attention to the interaction of time series, so as to better capture the correlation between time series and effectively encode such correlation into the realization of sequence characterization. The TGMformer structure with attenuating mask complies with the future-independent rule which is the basic feature of temporal data, and can better capture the potential patterns influenced by time steps.

\noindent\textbf{Industry graph group} In this task, we compare our TCGPN with other Spatial-Temporal Graph Neural Networks which allowed inputting the adjacency matrix with additional information. As shown in Table \ref{tab:result indus group}, it can be observed that our TCGPN method achieves the best performance in the majority of metrics, and we can clearly see a significant improvement compared to the baseline in fields such as IC, Calmar, Sharpe and more. This demonstrates the powerful ability of our TCGPN to integrate correlation information. By combining GAT with the correlation pre-training task, our model can effectively utilize graph information and combine it with temporal features more efficiently through attention mechanisms. As a result, our model exhibits greater improvement compared to other models. On the other hand, it is evident that STGNNs fail to effectively extract correlation features, and the introduction of additional information only leads to limited performance improvement, further highlighting the effectiveness of our method.

In summary, the experiments demonstrate that the combination of GAT and TGMformer modules effectively integrates temporal and correlation features, resulting in superior encoding. The setting of temporal and correlation pre-training tasks enables the model to have higher generalization and perform better on stock time series data without periodicity or fixed patterns.
\begin{figure}[htbp]
    \centering
    \includegraphics[width=\linewidth]{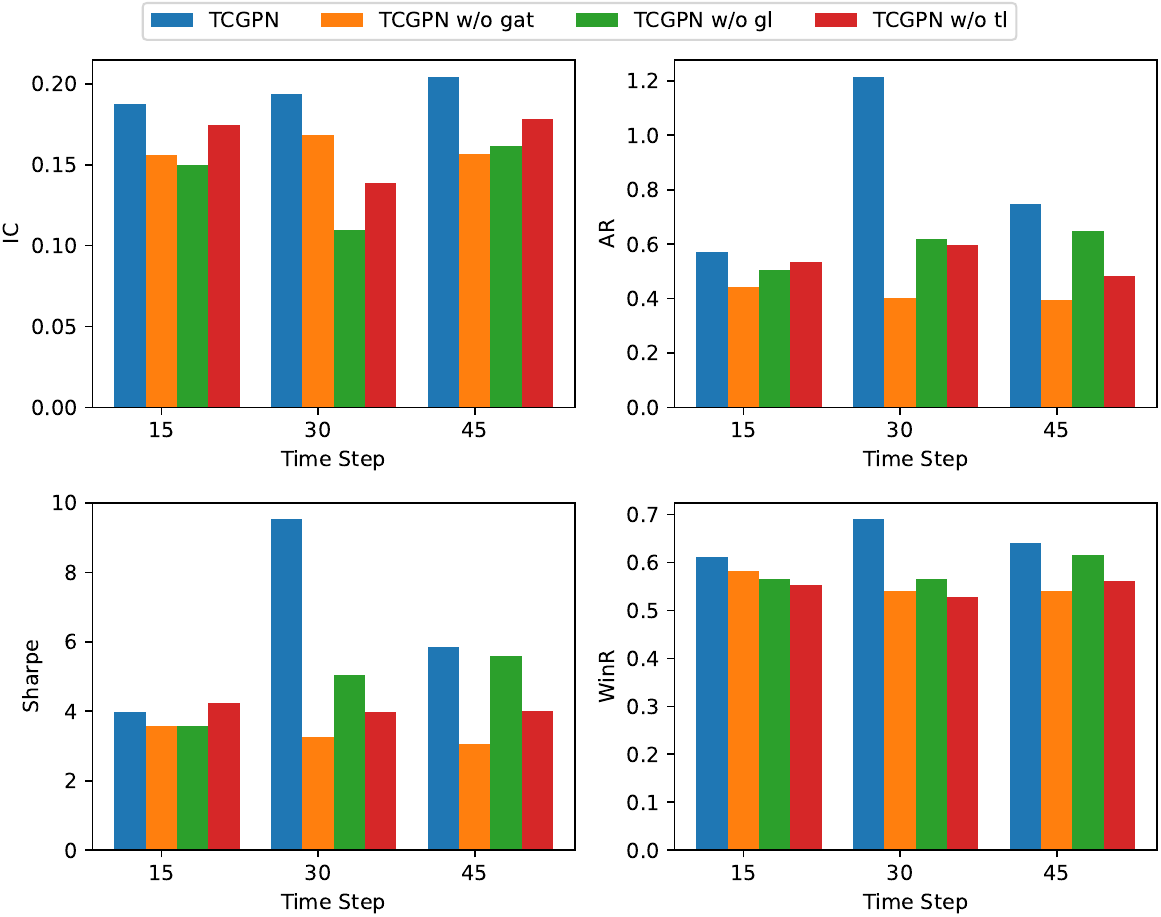}
    \caption[width=\textwidth]{Ablation study}
    \label{fig:ablation}
\end{figure}

\subsection{Ablation Study}
To validate the effectiveness of the key components, we conduct a ablation study on CSI500\&Industry Graph. We name variants of TCGPN as follows:
\begin{itemize}
\item \textbf{TCGPN w/o gat} Removing the Graph Attention Network from the model and solely utilize the TGMformer to model and restore temporal and correlation features.
\item \textbf{TCGPN w/o gl} Removing the correlation pre-training task from the model and exclude the calculation of graph loss. Instead, solely utilize time series pre-training task as the target.
\item \textbf{TCGPN w/o tl} Removing the time series pre-training task from the model and exclude the calculation of temporal loss. Instead, solely utilize correlation pre-training task as the target.
\end{itemize}

We repeated each experiment five times across three time steps \{15,30,45\} and plotted the results of representative metrics on Figure \ref{fig:ablation}. As can be seen from the Figure \ref{fig:ablation}, TCGPN outperforms any variant in terms of performance. Specifically, the decrease in IC indicates a decrease in the overall accuracy of predicting a single time step. Removing GAT or the correlation pre-training task from the model leads to a greater overall decrease in accuracy, indicating that both modules are necessary and mutually enhancing. Together, they effectively exploit correlation features. The relatively minor impact of removing time series pre-training task suggests that the learning of correlation features plays a role in prediction that is similar to or even greater than the temporal feature. For a local perspective, the impact of time series pre-training task increases with the growth of time steps, indicating time series pre-training task is important to model temporal features. Compared to removing correlation pre-training task, removing GAT leads to a greater decrease in model performance, suggesting that GAT can effectively learn the correlations between time series. On the other hand, correlation pre-training task enhances the robustness of the model, resulting in better overall performance.


\begin{figure}[htbp]
    \centering
    \includegraphics[width=\linewidth]{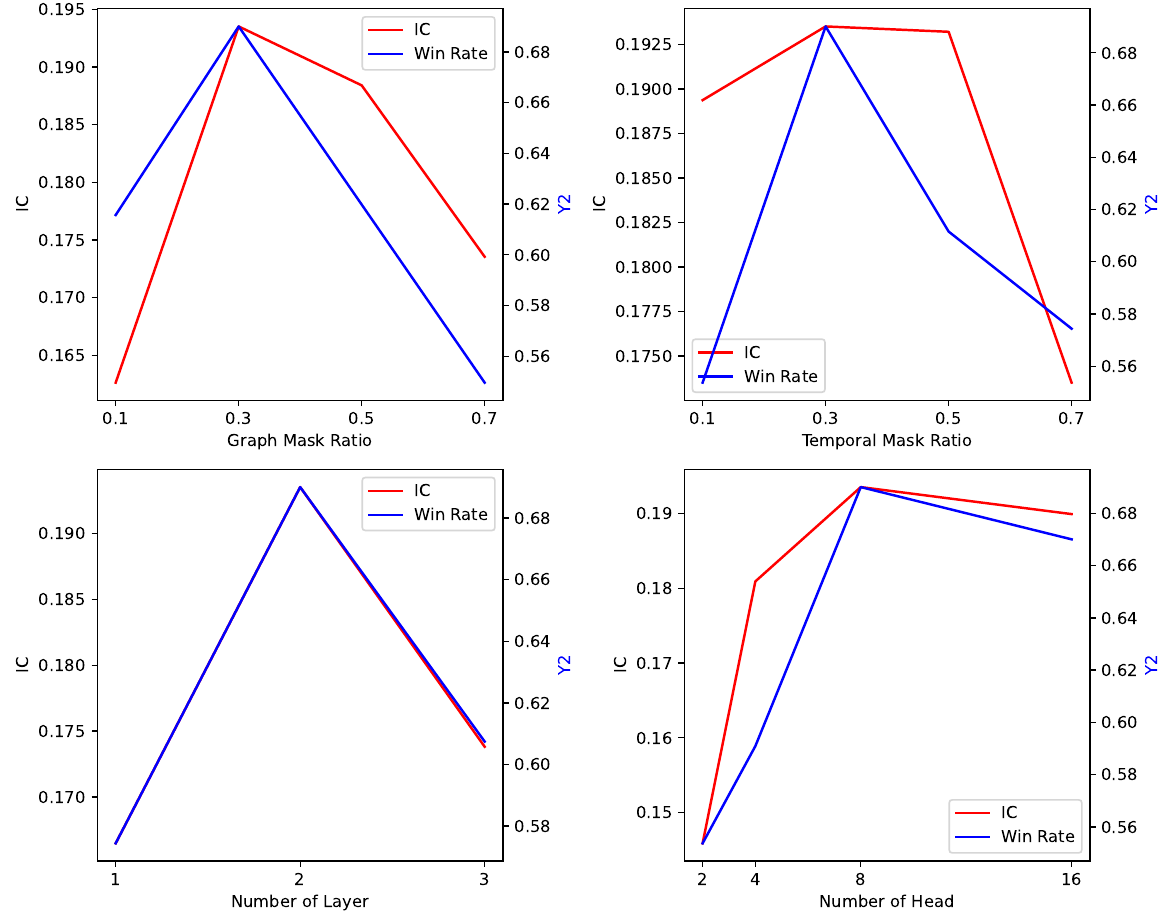}
    \caption[width=\textwidth]{Hyper-parameter study. Compare the changes of IC (left axis) and WinR (right axis).}
    \label{fig:hyper-param}
\end{figure}
\subsection{Hyper-parameter Study}
We conduct a large number of parameter experiments. Taking into account the model structure and pretraining tasks, we primarily investigate the impact of four hyperparameters on model performance: graph mask ratio $r_g$, temporal mask ratio $r_{t}$, number of layers $n_{l}$, and number of heads $n_{h}$. 
We perform repeated experiments on the CSI500\&Industry Graph dataset and used ic and win rate as evaluation metrics. The results are shown in the Figure \ref{fig:hyper-param}.

We find that both $r_{g}$ and $r_{t}$ reach their optimal values at around 0.3, but the reasons for their optimal values are different based on our analysis. When $r_{t}$ is below 0.3, the training task becomes too easy, and missing values can be filled through interpolation. When $r_{t}$ is 0.5 or higher, for time series of length 30, there is insufficient visible information to capture meaningful patterns of change. As for $r_{g}$, when it is below 0.3, the adjacency matrix computed by GAT is denser, leading to excessive ineffective connections and insufficient robustness. However, when $r_{g}$ is above 0.5, the computed matrix becomes too sparse, resulting in insufficient relevant information.

For $n_{l}$, it determines the complexity of temporal modeling. When the $n_{l}$ is 1, the model lacks encoding capacity, resulting in poor performance. When the model has 3 layers, it exhibits more severe overfitting issues for shorter time series and the model loses necessary robustness. Regarding $n_{h}$, which affects correlation modeling, the model achieves optimal performance when $n_{h}$ is 8. When $n_{h}$ is less than 8, the model fails to effectively capture multiple correlation relationships. As $n_{h}$ continues to increase, the model cannot explore additional correlation relationships and instead experiences a slight decrease in performance due to redundancy.

\section{RELATED WORK}
\subsection{Pre-trained Model}
Pre-training is one of the latest and most popular training methods
. Many existing pretrained models, such as BERT \cite{BERT} and GPT \cite{GPT}, utilize the Encoder and Decoder structures of Transformers to learn the correlations within sequences and obtain better encoding representations. These representations possess stronger robustness, richer information, and lower noise. 

In recent research, there are studies on enhancing downstream prediction tasks with pre-training on time series data, such as STEP \cite{STEP} and SPGCL \cite{SPGCL}. 
However, they only result in improved temporal representations without considering spatial correlations.

\subsection{Spatial-Temporal Graph Neural Network}
In recent years, there has been a growing interest in exploring the combination of temporal and spatio information within time series data
. This led to the emergence of a series of models known as Spatial-Temporal Graph Neural Networks.

STGCN \cite{STGCN} is the first to introduce Graph Convolutional Neural Network (GCN) \cite{GCN} into time series forecasting. 
Subsequently, more convolution-based models such as Graph Wave Net\cite{Gwave}, MTGNN \cite{MTGNN}, StemGNN \cite{stemGNN}, H-STGCN \cite{H-STGCN}, GSTNet \cite{GSTNet}, and others are proposed. These models introduce various gating mechanisms on top of convolutions to better capture data features. Meanwhile, some studies focus on more complex convolutional structures, such as ST-GDN \cite{ST-GDN} and ST-ResNet \cite{ST-ResNet}, which achieve better performance through clever architectural designs and mechanisms.

Furthermore, some works, like ARGCN \cite{ARGCN}, DCRNN \cite{DCRNN}, TGCN \cite{TGCN}, combine Recurrent Neural Networks (RNNs) with Graph Neural Networks (GNNs). They leverage the excellent temporal modeling capabilities of RNNs to better capture temporal features. In addition, with the introduction of Transformers, many models incorporate transformer architectures or attention mechanisms into spatial-temporal modeling, such as ASTGCN \cite{ASTGCN}, STGNN \cite{sTGNN}, and GMAN \cite{GMAN}. 

Although an increasing number of models are proposed, 

they perform well in tasks with periodic patterns but lack robustness in tasks without periodicity. Our approach can generate better representations in both temporal and spatial domains, exhibiting stronger robustness and generalization capabilities.

\section{CONCLUSION}
In this paper, we propose a novel Temporal-Correlation Graph Pre-trained Network called TCGPN. The temporal-correlation fusion encoder effectively integrates temporal information and correlation features. The temporal self-supervised task prompts the model to explore potential temporal contextual relationships and latent influence patterns between sequences. And the correlation semi-supervised task allows the model to uncover more effective relational linkages between sequences under the guidance of prior knowledge, enhancing the robustness and generalizability of the encoding. Additionally, TCGPN is independent of the number of nodes and their sequence. It can enhance the pre-training effect through various data augmentation methods and address the issue of excessive memory usage during spatio-temporal pre-training by utilizing repeated sampling techniques. We conduct extensive experiments on real-world stock datasets to demonstrate the superiority of our approach, and it is the first application of this method to stock market data where robustness is crucial. In the future, we will continue to explore better pre-training tasks and model structures, and apply our approach to a wider range of spatial-temporal tasks.

\begin{ack}
This work is supported by the National Natural Science Foundation of China (Grant No. 62276008, 62250037, and 62076010), and partially supported by the National Key R\&D of China (Grant \#2022YFF0800601).
\end{ack}

\bibliography{mybib}

\end{document}